\pdfoutput=1
\documentclass{article}

 \PassOptionsToPackage{numbers, compress}{natbib}

 \usepackage[preprint,nonatbib]{nips_2018}

\usepackage[utf8]{inputenc} 
\usepackage[T1]{fontenc}    

\usepackage{url}            
\usepackage{booktabs}       
\usepackage{amsfonts}       
\usepackage{nicefrac}       
\usepackage{microtype}      
\usepackage{hyperref}       

\usepackage[caption=false]{subfig}
\usepackage{bigints}

\usepackage{amsmath,amssymb}
\usepackage{amsthm}

\usepackage{algorithm}
\usepackage{algorithmic}
\usepackage{tikz}
\usepackage{multirow}
\usetikzlibrary{automata, positioning}

\title{GAN Q-learning}

\author{
  Thang Doan\thanks{These authors contributed equally.} \\
  Desautels Faculty of Management\\
  McGill University\\
  \texttt{thang.doan@mail.mcgill.ca} \\
  \And
  Bogdan Mazoure$^*$ \\
  Department of Mathematics \& Statistics \\
  McGill University\\
  \texttt{bogdan.mazoure@mail.mcgill.ca} \\
  \AND
  Clare Lyle \\
  School of Computer Science\\
  McGill University \\
  \texttt{clare.lyle@mail.mcgill.ca} \\
}
\begin{document}

\maketitle

\begin{abstract}
  Distributional reinforcement learning (distributional RL) has seen empirical success in complex Markov Decision Processes (MDPs) in the setting of nonlinear function approximation. However there are many different ways in which one can leverage the distributional approach to reinforcement learning. In this paper, we propose GAN Q-learning, a novel distributional RL method based on generative adversarial networks (GANs) and analyze its performance in simple tabular environments, as well as OpenAI Gym. We empirically show that our algorithm leverages the flexibility and blackbox approach of deep learning models while providing a viable alternative to traditional methods.
\end{abstract}

\section{Introduction}

Reinforcement learning (RL) has recently had great success in solving complex tasks with continuous control~\cite{trpo,ppo}. However, as these methods often have high variance results while dealing with unstable environments, distributional perspectives on the state-value function in RL have begun to gain popularity~\cite{bellemare2017distributional}. Note that the distributional perspective is distinct from Bayesian approaches to the RL problem as the former models the inherent variability of the returns from a state and not the agent's confidence in its prediction of the average return. 

Up to now, deep learning methods in RL used multiple function approximators (typically a network with shared hidden layers) to fit a state value or state-action value distribution. For instance, \cite{bootstrappedDQN} used $k$-heads on the state-action value function $Q$ for every available action and used it to model a distribution. 
In \cite{bayesianpol}, a Bayesian framework was applied to the actor-critic architecture by fitting a Gaussian Process (GP) instead of the critic, hence allowing for a closed-form derivation of update rules. More recently, \cite{bellemare2017distributional} introduced a distributional algorithm C51 which aimed to solve the RL problem by learning a categorical probability vector over returns $Q$. Unlike \cite{GANRL} which uses a generative network to learn the underlying transition model of the environment, we utilize a generative network to model the distribution approximation of the Bellman updates.\\
In this work, we build on top of the aforementioned distributional RL methods and introduce a novel way to learn the state-value distribution. Inspired by the analogy between the actor-critic architecture and generative adversarial networks (GANs) \cite{connection_gan_actor_critic}, we leverage the later in order to implicitly represent the distribution of the Bellman target update through a generator/discriminator architecture. We show that, although sometimes volatile, our proposed algorithm is a viable alternative to now considered classical deep Q networks (DQN). We aim to provide a unifying view on distributional RL through the minimization of the Earth-Mover distance and without explicitly using the distributional Bellman projection operator on the support of the $Q$ values.

\section{Related Work}
\subsection{Background}

Multiple tasks in machine learning require finding an optimal behaviour in a given setting, i.e. solving the reinforcement learning problem. We proceed to formulate the task as follows.\\
Let ($S$,$A$,$R$,$P$,$s_0$,$\gamma$) be a 5-tuple representing a \textit{Markov decision process} (MDP) where $S$ is the set of states, $A$ the set of allowed actions, $R:S\times A \rightarrow \mathbb{R}$ is the (deterministic or stochastic) reward function, $P$ are the environment transition probabilities and $s_0 \subseteq S$ is the set of initial states. At a given time step $t=1,2,3,...$ an agent acts according to a policy $\pi(a|s)=P[a_t|s_t]$. The environment is characterized by its set of initial states in which the agent starts, as well as the transition model which encodes the mechanics of moving from one state to another. In order to compare states, we introduce the state value function $V:S \to \mathbb{R}$ which gives the expected sum of discounted rewards in that state. That is, $V(s)=\mathbb{E}_{\pi, P}[\sum_{t=1}^\infty\gamma^tR_t|S_0=s]$.\\
The reinforcement learning problem is two-fold: (1) given a fixed policy $\pi$ we would like to obtain the correct state value function $V_\pi(s), \forall s \in S$ and (2) we wish to find the optimal policy $\pi^*(a)$ which yields the highest $V(s)$ for all states of the MDP or, equivalently, $\pi^*=\text{argmax}_{\pi}V_\pi(s)$. The first task is known in the reinforcement learning literature as prediction and the second as control.\\
In order to find the value function for each state, we need to solve the Bellman equations \cite{bellman1954theory}:
\begin{equation}
    \begin{split}
        V_\pi(s)&=\mathbb{E}_{\pi,P}[R(s_{t},a_{t})+\gamma V_\pi(s_{t+1})|s_t = s],
    \end{split}
    \label{eq:bellman_v_eq}
\end{equation}
for $V_\pi=\{V_\pi(s_1),...,V_\pi(s_n)\}$ and $R=\{R(s_1,a_1),...,R(s_n,a_n)\}$. If we define a state-action value function $Q_\pi(s,a):S \times A \to \mathbb{R}$ as $Q_\pi(s,a)=\mathbb{E}_{\pi, P}[\sum_{t=0}^\infty \gamma^t R(S_t,A_t) | S_0 = s, A_0 = a]$, then we can rewrite Eq.\ref{eq:bellman_v_eq} as:
\begin{equation}
    \begin{split}
    Q_\pi(s,a)&=\mathbb{E}_{\pi,P}[R(s,a)+\gamma Q_\pi(s',a')] \text{ or }\\
    Q_\pi&=R+\gamma P_\pi Q_\pi \;,
    \end{split}
    \label{eq:bellman_q_eq}
\end{equation}
for $Q_\pi=\{Q_\pi(s_1,a_1),...,Q_\pi(s_n,a_n)\}$.\\
While both Eq.\ref{eq:bellman_v_eq} and Eq.\ref{eq:bellman_q_eq} have the following direct solution obtained by matrix inversion: $V_\pi=(I-\gamma P)^{-1}R$. However it should be noted that this is only well-defined for finite-state MDPs, and further as this computation has $O(n^{2.4})$ complexity \cite{coppersmith1987matrix}, it is only computationally feasible for small MDPs. Therefore, sample-based algorithms such as Temporal Difference (TD) \cite{sutton1988learning} for prediction and SARSA or Q-learning \cite{rummery1994line,watkins1992q} for control are preferred for more general classes of environments.

\subsection{Distributional Reinforcement Learning}

In the setting of distributional reinforcement learning \cite{bellemare2017distributional}, we seek to learn the distribution of returns from a state, rather than the mean of the returns. We translate the Bellman operator on points to an operator on distributions. The vector of mean rewards therefore becomes a function of reward distributions $R(s,a)$. We can thus represent $Q_\pi(s,a)$ as $Z_\pi(s,a)$, a random variable whose law is the returns following a state. Both expected and distributional distributional quantities are linked through the following relation: $Q_\pi(s,a)=\mathbb{E}[Z_\pi(s,a)]$. We finally arrive to the \textit{distributional Bellman equations}:
\begin{equation}
\begin{split}
\mathcal{T}^\pi Z_\pi(s,a) &\stackrel{d}{\equiv} R(s,a) + \gamma Z_\pi(S',A') \text{ with fixed point: }\\
     Z_\pi(s,a) &\stackrel{d}{=} \mathcal{T}^\pi Z_\pi(s,a), \forall s \in S \; \forall a \in A
    \label{eq:distributional_bellman}
\end{split}
\end{equation}
where $\mathcal{T}^\pi$ denotes an analogue to the well-known Bellman operator now defined over distributions.\\
Eq.\ref{eq:distributional_bellman} is the distributional counterpart of Eq.\ref{eq:bellman_q_eq}, where equality holds for sequences of random variables.

In traditional reinforcement learning algorithms, we use experience in the MDP to improve an estimated state value function in order to minimize the expected distance between the value function's output for a state and the actual returns from that state. In distributional reinforcement learning, we still aim to minimize the distance between our output and the true distribution, but now we have more freedom in how we choose our definition of "distance", as there are many metrics on probability distributions which have subtly different properties. 

The $p-$Wasserstein metric between two real-valued random variables $X$ and $Y$ with cumulative distribution functions $F_X$ and $F_Y$ is given by
\begin{equation}
    W_p(F_X,F_Y)=\bigg(\inf_{(X,Y)} \mathop{\mathbb{E}}_{(X,Y)}\big[|X-Y|^p\big]\bigg)^{\frac{1}{p}}
\end{equation}
For $p=1$ we recover the widely known Earth-Mover distance. More generally, for any $1 \leq p \leq q \leq \infty$, $W_p(F_X,F_Y)\leq W_q(F_X,F_Y)$ holds \cite{givens1984class} and is useful in contraction arguments. The maximal $p-$Wasserstein metric, $\overline{W}_p=\sup_{(s,a)} W_p(Z(s,a),Z'(s,a))$ is defined over state-action tuples $(s,a) \in S \times A$  for any two value distributions $Z,Z'$. 

It has been shown that while $\mathcal{T}^{\pi}$ is a contraction under the maximal $p-$Wasserstein metric \cite{ruschendorf1985wasserstein}, the Bellman \textit{optimality} operator $\mathcal{T}$ is not necessarily a contraction. This result implies that the control setting requires a treatment different from prediction, which is done in \cite{bellemare2017distributional} through the C51 algorithm in order to guarantee proper convergence.

\subsection{Generative Models}

Generative models such as hidden Markov models (HMMs), restricted Boltzmann machines (RBMs) \cite{salakhutdinov2007restricted}, variational auto-encoders (VAEs) \cite{kingma2013auto} and generative adversarial networks (GANs) \cite{goodfellow2014generative} learn the distribution of data for all classes. Moreover, they provide a mechanism which allows to sample new observations from the learned distribution.\\
In this work, we make use of generative adversarial networks which consist of two neural networks playing a zero-sum game against each other. The generator network $G:Z \to X$ is a mapping from a high-dimensional noise space $Z = \mathbb{R}^{d_z}$ onto the input space $X$ on which a target distribution $f_X$ is defined. The generator's task consists in fitting the underlying distribution of observed data $f_X$ as closely as possible. The discriminator network $D:X \to \mathbb{R} \cap [0,1]$ scores each input as the probability of coming from the real data distribution $f_X$ or from the generator $G$. Both networks are gradually improved through alternating or simultaneous gradient descent updates.

The classical GAN algorithm minimizes the Jensen-Shannon divergence (JS) between the real and generated distributions.
Recently, \cite{arjovsky2017wasserstein} suggested to replace the JS metric by the Wasserstein-1 or Earth-Mover divergence. We make use of an improved version of this algorithm, Wasserstein GAN with Gradient Penalty \cite{gulrajani2017improved}. It's objective is given below:
\begin{equation}
    \min_G\max_D \underset{{x \sim f_X(x)}}{\mathbb{E}}[D(x)]+\underset{{x \sim G(z)}}{\mathbb{E}}[-D(x)]+p(\lambda),
    \label{eq:wass_gan_loss}
\end{equation}
where $p(\lambda)=\lambda(||\nabla_{\tilde{x}} D(\tilde{x})||-1)^2,\; \tilde{x}=\varepsilon x + (1-\varepsilon)G(Z)$,  $\varepsilon \sim \text{Uniform}(0,1)$, and $Z\sim f_Z(z)$. Setting $\lambda=0$ recovers the original WGAN objective.\\




\section{GAN Q-learning}
\subsection{Motivation}

We borrow the two-player game analogy from the original GAN paper \cite{goodfellow2014generative}: the generator network's purpose is to produce realistic samples of the optimal state-action value distribution (estimate of $Z_{\pi^*}(s,a)$). On the other hand, the discriminator network aims to distinguish real samples of $\mathcal{T} Z(s,a)$ from the samples $Z(s,a)$ outputted by $G$.
The generator network improves its performance through the signal from the discriminator, which is reminiscent of the actor-critic architecture  \cite{connection_gan_actor_critic}.


\subsection{Algorithm}

At each timestep, $G$ receives stochastic noise $z \sim \mathcal{N}(0,1)$ and a state $s$ as input and returns a sample $G(z|(s,a))$ for every action $a$ from the current estimate  of $Z(s,a)$. We then select the action $a^* =\underset{a}{\max}\; G(z|(s,a))$. The agent then applies the chosen action $a^*$, receives a reward $r$ and transitions to state $s'$. The tuple $(s,a,r,s')$ is saved into a replay buffer $\mathcal{B}$ as done in \cite{dqn}. Each update step consists in sampling a tuple $(s,a,r,s')$ uniformly from the buffer $\mathcal{B}$ and proceed to update the generator and discriminator according to Eq~\ref{eq:wass_gan_loss}. 

Values obtained from the Bellman backup operator $x\equiv x(s,a,r,s')= r+\gamma \underset{a}{\max}G(z|(s',a))$ are considered as coming from the real distribution. The discriminator's goal is to differentiate between the Bellman target $x=r+\gamma \underset{a}{\max} G(z|(s',a))$ and the output produced by $Z(s,a)$. We obtain the following updates for $G$ and $D$, respectively:
\begin{equation}
\mathcal{L}(w_D,w_G)=\begin{cases}
\underset{(s,a,r,s') \sim \mathcal{B}}{\mathbb{E}}[D_{w_D}(x|(s,a))]-\underset{{(s,a) \sim \mathcal{B}}\atop {{z \sim \mathcal{N}(0,1)}\atop {X\sim G_{w_G}(z|(s,a))}}}{\mathbb{E}}[D_{w_D}(X|(s,a))]+p(\lambda)\;, & \\
\underset{{(s,a) \sim \mathcal{B}}\atop {{z \sim \mathcal{N}(0,1)}\atop {X\sim G_{w_G}(z|(s,a))}}}{\mathbb{E}}[-D_{w_D}(X|(s,a))]\;, &
\end{cases}
\end{equation}
where $w_G,w_D$ are weights of the generator and discriminator networks with respect to which the gradient updates $w^{(t+1)} \leftarrow w^{(t)}-\alpha_t\nabla_{w^{(t)}} \mathcal{L}(w^{(t)})$ are taken.

Note that to further stabilize the training process, one can use a target network $G'$ updated every $k$ epochs as in \cite{dqn}. Due to the nature of GANs, we advise training the model in a batch setting using experience replay and a second (target) network for a more stable training process.

\begin{algorithm}[H]
   \caption{GAN Q-learning}
   \label{alg:dist_gan}
\begin{algorithmic}
   \STATE {\bfseries Input:} MDP $M$, discriminator and generator networks $D,G$, learning rate $\alpha$,$n_{disc}$ the number of updates of the discriminator, $n_{gen}$ the number of updates of the generator, gradient penalty coefficient $\lambda$, batch size $m$.\\
   
   \STATE Initialize replay buffer $\mathcal{B}$ to capacity N, $D$ and $G$ with random weights, $Z$, $s_0, a_0$.
   \STATE $t \gets 0$
   \FOR{$\text{episode}=1,...,M$} 
   \FOR{$t=1,...,T_{max}$}
   \STATE sample a batch $z \sim N(0,1)$
   \STATE $a_t \leftarrow \underset{a}{\max} G(z|(s_t,a))$
   \STATE sample $s_{t+1} \sim P(\cdot|s_t,a_t)$
   \STATE Store transition ($s_t$,$a_t$,$r_t$,$s_{t+1}$) in $\mathcal{B}$
   \STATE \COMMENT{Updating Discriminator}
   \FOR{$n=1,...,n_{disc}$}
    \STATE Sample minibatch $\{ s,a,r,s' \}_{i=1}^{m}$ from $\mathcal{B}$
   \STATE sample batch $\{z \}_{i=1}^{m} \sim N(0,1)$

   \STATE Set $y_{i}= \begin{cases}
   r_{i}, & \text{$s^\prime$ terminal} \\
    r_{i} + \gamma \underset{a_{i}}{\max}\; G(z_{i}|(s_{i}^\prime,a_{i})), & \text{otherwise}\\
    \end{cases}$
    
   \STATE sample a batch $\{ \epsilon \}_{i=1}^{m} \sim N(0,1)$
   \STATE $\tilde{x_{i}}\leftarrow\epsilon_{i} y_{i}+(1-\epsilon_{i})\underset{a_{i}}{\max}\; G(z_{i}|(s_{i}',a_{i}))$
   \STATE $  \mathcal{L}^{(i)}  \gets   D(G(z_{i}|(s_{i},a_{i}^*)|(s_{i},a_{i}^*))-D(y_{i}|(s_{i},a_{i}^*))+\lambda (| \nabla_{\tilde{x}} D(\tilde{x_{i}}|(s_{i},a_{i}^*)) |-1)^{2}$

   $w_{D} \gets$ RMSProp($\nabla_{w_{D}}\frac{1}{m}\sum_{i=1}^{m} \mathcal{L}^{(i)},\alpha $ )
   \ENDFOR
   \STATE \COMMENT{Updating Generator}
   \FOR{$n=1,...,n_{gen}$}
   \STATE sample a batch of $\{z^{(i)}\}_{i=1}^{m} \sim N(0,1)$
   \STATE $w_{G} \gets$ RMSProp($-\nabla_{w_{G}}\frac{1}{m}\sum_{i=1}^{m}\mathcal{L}^{(i)},\alpha   $ )
     \ENDFOR
    \ENDFOR
    \ENDFOR
    
\end{algorithmic}
\end{algorithm}
 
Here, the objective function $\mathcal{L}$ is identical to Eq.\ref{eq:wass_gan_loss}, where $f_X(x)$ is taken to be $r(s,a)+\gamma\underset{a}{\max}\; G(z|(s',a))$.

\begin{figure}
    \centering
    \includegraphics[width=0.7\linewidth]{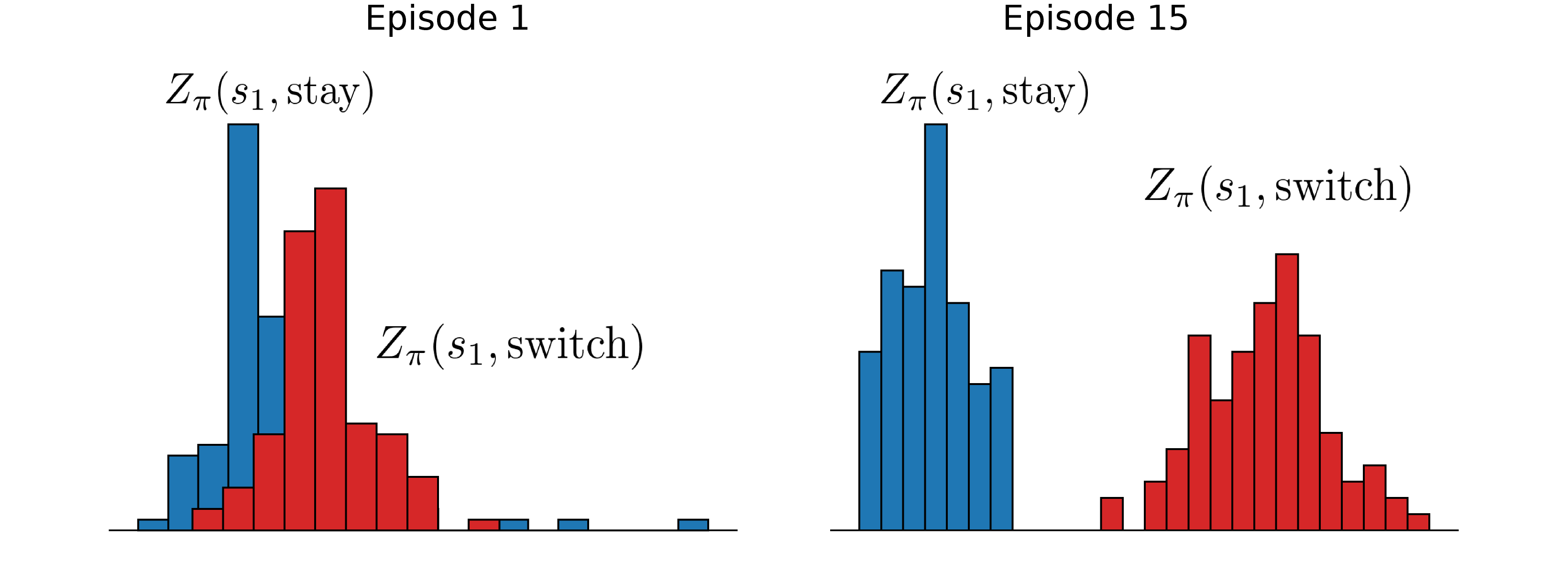}
    \caption{Evolution of state value distribution learned by GAN Q-learning on the Two State environment. Here, $\pi$ denotes the greedy policy.}
    \label{fig:distribution_split}
\end{figure}

Using a generative model to represent the state-action value distribution allows for an alternative to explicit exploration strategies. Indeed, at the beginning of the training process, taking an action is analogous to using a decaying exploration strategy (since $G$ has not clearly separated $Z(s,a)$ for every $(s,a)$ pair). Fig.~\ref{fig:distribution_split} demonstrates how gradually separating $Z(s,a)$ from $\mathcal{T} Z(s,a)$ acts as implicit exploration by sampling suboptimal actions.

\section{Convergence}
It is well-known that Q-learning can exhibit divergent behaviour in the case of nonlinear function approximation \cite{tsitsiklis1997analysis}. Because nonlinear value function approximation can be viewed as a special case of the GAN framework where the latent random variable is sampled from a degenerate distribution, we can see that the class of problems for which GAN Q-learning can fail to converge contains those for which vanilla Q-learning does not converge. 

Further, as explored in \cite{mescheder2018convergence}, we observe that in many popular GAN architectures, convergence to the target distribution is not guaranteed, and oscillatory behaviour can be observed. This is a double blow to the reinforcement learning setting, as we must guarantee both that a stationary distribution exists and that the GAN architecture can converge to this stationary distribution.

We also note that although in an idealized setting for the Wasserstein GAN the generator should be able to represent the target distribution and the discriminator should be able to learn any 1-Lipschitz function in order to produce the true Wasserstein distance, this is unlikely to be the case in practice. It is thus possible for the optimal generator-discriminator equilibrium to correspond to a generated distribution that has a different expected value from the target distribution. Consider, for example, a generator which produces a Dirac distribution, and a discriminator which can compute quadratic functions of the form $m_Dx^2$. Then the discriminator attempts to approximate the Wasserstein distance by computing
\[ \max_{m_D \in \mathbb{R}}\int_{-\infty}^\infty m_D x^2 d[f_G(x) - f_T(x)] \]

Suppose we are in a 2-armed bandit setting, where arm A always returns a reward of $\frac{1}{2} + \epsilon$ for some small $\epsilon>0$ and arm B gives rewards distributed as a $\text{Bernoulli}(1/2)$. Then the optimal generator (constrained to the class of Dirac distributions) will predict a Dirac distribution with support $\frac{1}{2} + \epsilon$ for arm A, and a Dirac distribution with support $\frac{1}{\sqrt{2}}$ for arm B. Consequently, an agent which has reached an equilibrium will incorrectly estimate arm B as being the optimal arm. 

Empirical results reported in the next section demonstrate the ability of our algorithm to solve complex reinforcement learning tasks. However, providing convergence results for nonlinear $G$ and $D$ can be hindered by complex environment dynamics and the unstable nature of GANs. For instance, proving convergence of the generator-discriminator tuple to a saddle point requires an argument similar to \cite{mescheder2018convergence}.

\section{Experiments} 
In order to compare the performance of the distributional approaches to traditional algorithms such as Q-learning, we conducted a series of experiments on tabular and continuous state space environments.

\subsection{Environments}

We considered the following environments:
\begin{enumerate}
    \item 10-state chain with two goal states (\textbf{2G Chain}) ($s_0$ is in the middle). Two deterministic actions (left, right) are allowed. A reward of $+1$ is given when we stay in the goal state for one step, $0$ otherwise. The discount factor $\gamma=0.6$ and the maximum episode length is 50;
    \item Deterministic $6 \times 6$ gridworld (\textbf{Gridworld}) with start and goal states in opposing corners and walls along the perimeter. A reward of 0 is given in the goal state, $-1$ otherwise. The agent must reach the goal tile in the least number of steps while avoiding being stuck against walls. The discount factor $\gamma=0.9$ and the maximum episode length is 100;
    \item The simple two state MDP (\textbf{2 States}) presented in Fig.\ref{fig:two_state_mdp}. The discount factor $\gamma=0.95$ and the maximum episode length is 25.
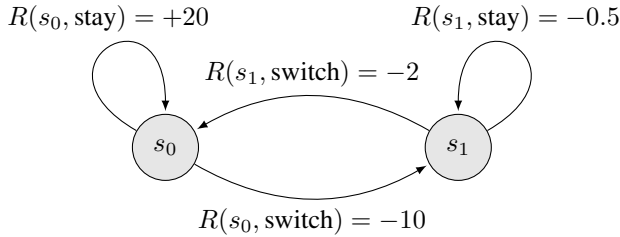
\begin{figure}[h]
    \centering
     \begin{tikzpicture}
        \tikzset{node style/.style={state, 
                                    fill=gray!20!white,
                                    circle}}
    
        \node[node style]               (I)   {$s_0$};
        \node[node style, right=3cm of I]   (II)  {$s_1$};

        \draw[>=latex,
              auto=left,
              every loop]
              (I)   edge[loop above,in=90,out=150,looseness=10] node[yshift=0.1cm] {$R(s_0,\text{stay})=+20$} (I)
             (I)   edge[bend right=30,auto=right] node {$R(s_0,\text{switch})=-10$} (II)
             (II)   edge[bend right=30,auto=right] node {$R(s_1,\text{switch})=-2$} (I)
             (II)   edge[loop above,in=90,out=30,looseness=10] node[yshift=0.1cm] {$R(s_1,\text{stay})=-0.5$} (II);
    \end{tikzpicture}
    \caption{Simple two-state MDP.}
    \label{fig:two_state_mdp}
\end{figure}
    \item OpenAI Gym \cite{brockman2016openai} environments \textbf{Cartpole-v0} and \textbf{Acrobot-v1}.
    \end{enumerate}

All experiments were conducted with a similar, one hidden layer architecture for GAN Q-learning and DQN. A total of 3 dense layers of 64 units for tabular and 128 units for OpenAI environments each, as well as $tanh$ nonlinearities were used. Note that a Convolutional Neural Network (CNN) \cite{krizhevsky2012imagenet} can be used to learn the rewards similarly to \cite{dqn}.
\subsection{Results}
In this section, we present empirical results obtained by our proposed algorithm. For shorthand notation we abbreviate the tabular version of distributional Q-learning to dQ-learning; GAN Q-learning will be contracted to GAN-DQN. The dQ-learning performs a mixture update between the predicted and target distribution for each state-action pair it visits, analogous to TD updates.\\
All tabular experiments were ran on 10 initial seeds and the reported scores are averaged across 300 episodes. Gym environments were tested on 5 initial seeds over 1000 episodes each with no restrictions on maximum number of steps.

\begin{figure}[h]
\centering
\includegraphics[width=0.9\linewidth]{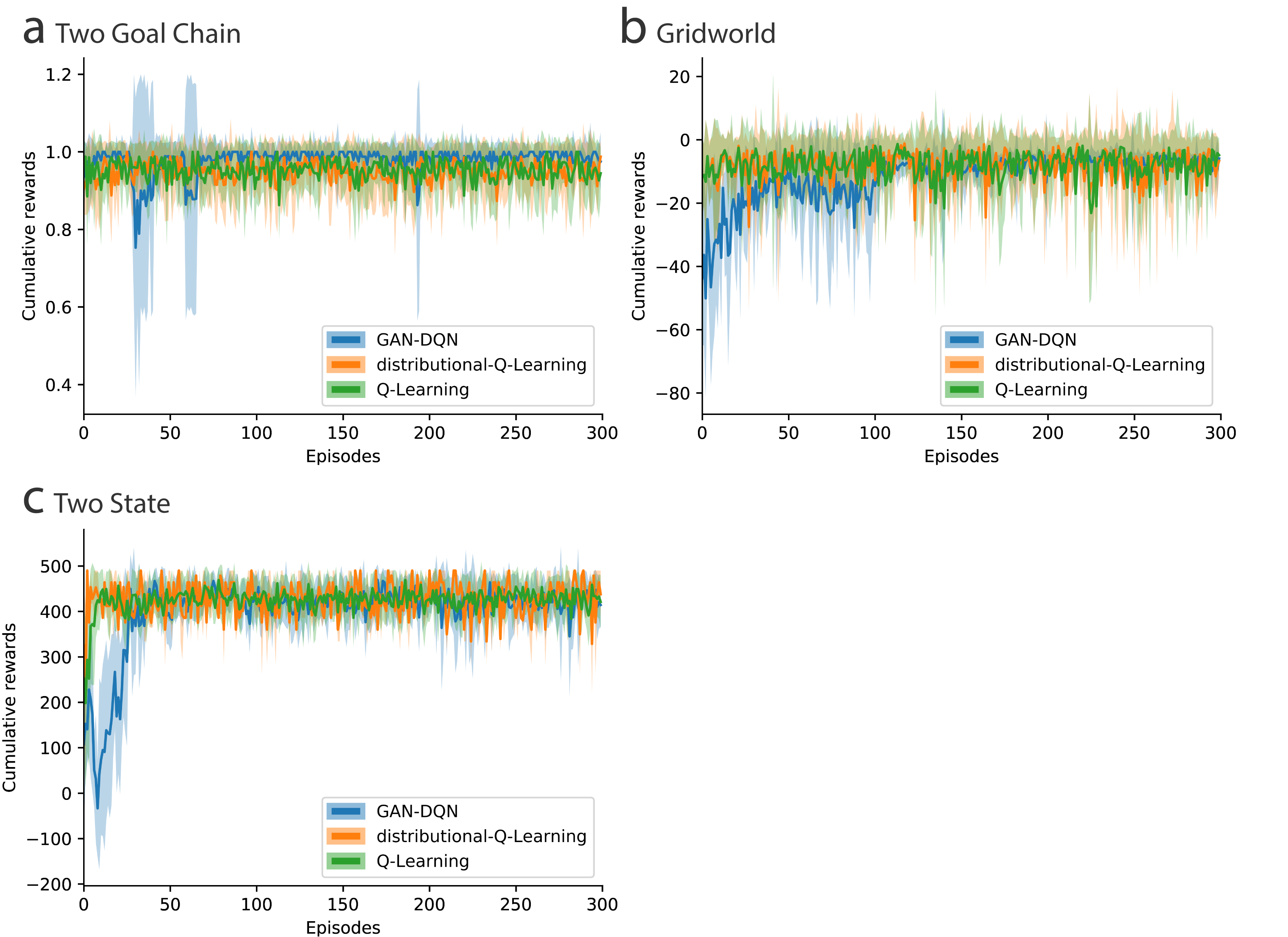}
\caption{Average cumulative rewards $r \pm \sigma_r$ obtained by distributional and expected agents in tabular environments.}
\label{fig:tabular_rewards}
\end{figure}

\begin{figure}[h]
    \centering
    \includegraphics[width=0.4\linewidth]{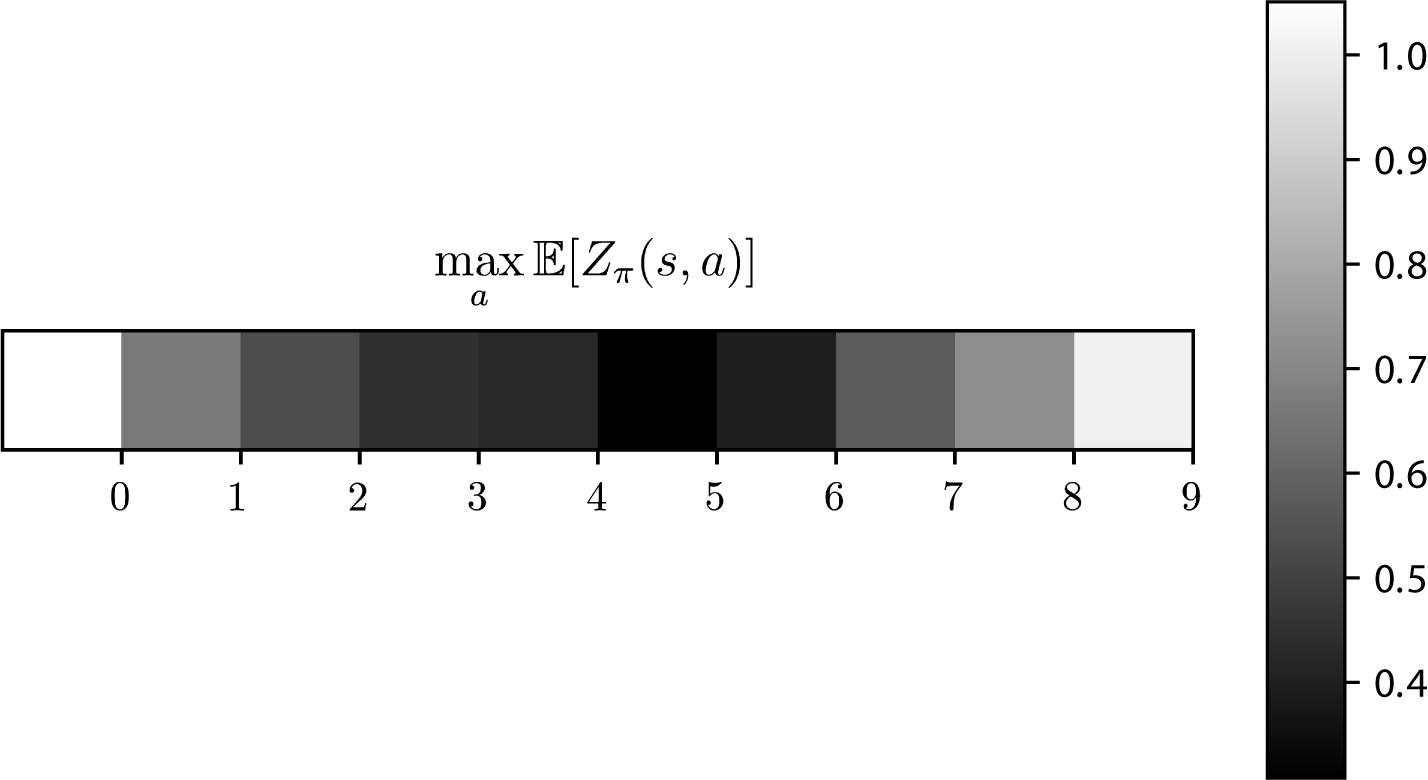}
    \caption{Expected state value function found by GAN Q-learning after 50 episodes in the Two Goal Chain environment for the greedy policy $\pi$.}
    \label{fig:2g_chain_state_value}
\end{figure}

In general, our GAN Q-learning results go on par with ones obtained by tabular baseline algorithms such as Q-learning (see Fig.~\ref{fig:tabular_rewards} and Table~\ref{table:average_steps_env_alg_ql}). The high variance in the first few episodes corresponds to the time needed for the generator network $G$ to separate real and generated distributions for each actions. Note that although the Gridworld environment yields sparse rewards similar to Acrobot, GAN Q-learning eventually finds the optimal path to the end goal. Fig.~\ref{fig:2g_chain_state_value} shows that our method can efficiently use the greedy policy in order to learn the state value function; lower state values are associated with the start state while the higher state values are attributed to both goal states.

\begin{figure}[H]
\centering
\includegraphics[width=0.9\linewidth]{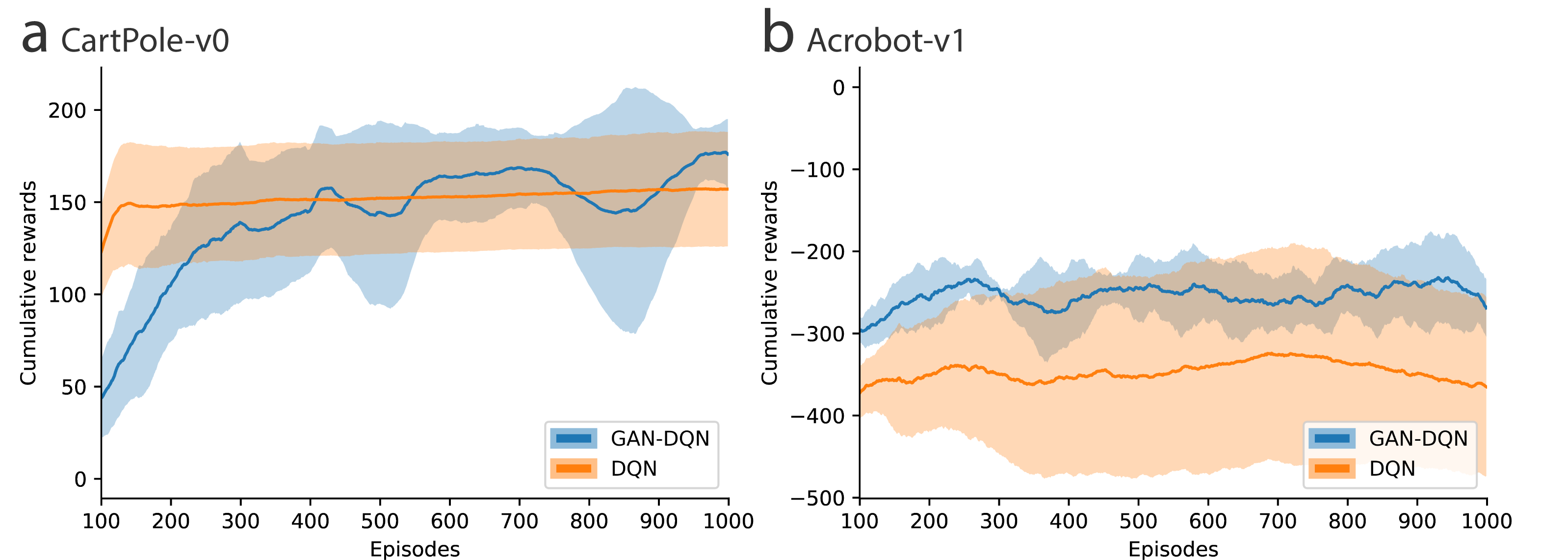}
\caption{Average cumulative rewards $r\pm \sigma_r$ obtained by distributional and expected agents in OpenAI Gym environments.}
\label{fig:openai_rewards}
\end{figure}

Just like DQN, the proposed method demonstrates the ability to learn an optimal policy in both OpenAI environments. For instance, increasing the number of updates for $G$ and $D$ in the \texttt{CartPole-v0} environment stabilized the algorithm and ensured the proper convergence of $G$ to the Bellman target. Fig.~\ref{fig:openai_rewards} demonstrates that, although sometimes unstable, control with GAN Q-learning has the capacity to learn an optimal policy and outperform DQN. Due to sparse rewards in \texttt{Acrobot-v1}, we had to rely on a target network $G'$ as in \cite{dqn} to stabilize the training process. Using GAN Q-learning without a second network in such environments would lead to increased variance in the agent's predictions and is hence discouraged.

\begin{table}[h]
\caption{Performance of Q-learning algorithms in tabular environments (in rewards/episode).}
\label{table:average_steps_env_alg_ql}
\begin{center}
\begin{small}
\begin{sc}
\begin{tabular}{lcccr}
& \multicolumn{3}{c}{Environment} \\
\hline
Algorithms & 2 State & 2G Chain & Gridworld\\
\hline
 Q-learning & 426.613 & 0.953 & \textbf{-8.059}\\
 dQ-learning & \textbf{427.328} & 0.950 & -8.190\\
 GAN-DQN & 398.918 & \textbf{0.978} & -11.720\\
\hline
\end{tabular}
\end{sc}
\end{small}
\end{center}
\end{table}

In addition to the common variance reduction practices mentioned above, we used a learning rate scheduler as a safeguard to reduce instability during the training process. We found that $\alpha_t=\frac{\alpha_0}{1+\frac{t}{k}}$ for timestep $t$, initial learning rate $\alpha_0$ and $k$ varying between environments (for \texttt{CartPole-v0}, $k=200$) yielded the best results. Even though our model has an implicit exploration strategy induced by the generator, we used an $\varepsilon-$greedy exploration policy like most traditional algorithms. We noticed that introducing $\varepsilon-$greedy the generator in separating the state value distributions for all actions. Note that for environments with less complex dynamics (e.g. tabular MDPs), our method does not require an explicit exploration strategy and has shown viable performance without it.

Unlike in the original WGAN-GP paper where a strong Lipschitz constraint is enforced via a gradient penalty coefficient ($\lambda$=10), we observed empirically that relaxing this property with $\lambda$=0.1 gave better results.

\section{Discussion}
We introduced a novel framework based on techniques borrowed from the deep learning literature in order to successfully learn the state-action value distribution via an actor-critic like architecture. Our experiments indicate that GAN Q-learning can be a viable alternative to classical algorithms such as DQN while having the appealing characteristics of a typical deep learning blackbox model. The parametrization of the returns distribution $Z(s,a)$ by a neural network within the scope of our approach is countered by its volatility in environments with particularly sparse rewards. We believe that a thorough understanding of the nonlinear dynamics of generative nets and convergence properties of MDPs is mandatory for a successful improvement of the algorithm. Recent work in the field hints that a saddle-point analysis of the objective function is a valid way to approach such problems \cite{mescheder2018convergence}.

Future work should address with high priority the stability of the training iteration. Moreover, using a CNN on top of screen frames in order to encode the state can provide a meaningful approximation to the reward distribution. Our proposed algorithm opens possibilities to integrate the GAN architecture into more complex algorithms such as DDPG \cite{ddpg} and TRPO \cite{trpo}, which can be a potential topic of investigation.

\section{Acknowledgments}
We would like to thank Marc G. Bellemare from Google Brain for helpful advice throughout this paper.

\bibliographystyle{plain}


\begin{thebibliography}{10}

\bibitem{arjovsky2017wasserstein}
Martin Arjovsky, Soumith Chintala, and L{\'e}on Bottou.
\newblock Wasserstein gan.
\newblock {\em arXiv preprint arXiv:1701.07875}, 2017.

\bibitem{bellemare2017distributional}
Marc~G. Bellemare, Will Dabney, and R{\'e}mi Munos.
\newblock A distributional perspective on reinforcement learning.
\newblock In {\em Proceedings of the 34th International Conference on Machine
  Learning}, 2017.

\bibitem{bellman1954theory}
Richard Bellman.
\newblock The theory of dynamic programming.
\newblock {\em Bulletin of the American Mathematical Society}, 60(6):503--515,
  1954.

\bibitem{brockman2016openai}
Greg Brockman, Vicki Cheung, Ludwig Pettersson, Jonas Schneider, John Schulman,
  Jie Tang, and Wojciech Zaremba.
\newblock Openai gym.
\newblock {\em arXiv preprint arXiv:1606.01540}, 2016.

\bibitem{coppersmith1987matrix}
Don Coppersmith and Shmuel Winograd.
\newblock Matrix multiplication via arithmetic progressions.
\newblock In {\em Proceedings of the nineteenth annual ACM symposium on Theory
  of computing}, pages 1--6. ACM, 1987.

\bibitem{bayesianpol}
Mohammad Ghavamzadeh, Yaakov Engel, and Michal Valko.
\newblock Bayesian policy gradient and actor-critic algorithms.
\newblock {\em Journal of Machine Learning Research}, 17(66):1--53, 2016.

\bibitem{givens1984class}
Clark~R Givens, Rae~Michael Shortt, et~al.
\newblock A class of wasserstein metrics for probability distributions.
\newblock {\em The Michigan Mathematical Journal}, 31(2):231--240, 1984.

\bibitem{goodfellow2014generative}
Ian Goodfellow, Jean Pouget-Abadie, Mehdi Mirza, Bing Xu, David Warde-Farley,
  Sherjil Ozair, Aaron Courville, and Yoshua Bengio.
\newblock Generative adversarial nets.
\newblock In {\em Advances in neural information processing systems}, pages
  2672--2680, 2014.

\bibitem{gulrajani2017improved}
Ishaan Gulrajani, Faruk Ahmed, Martin Arjovsky, Vincent Dumoulin, and Aaron~C
  Courville.
\newblock Improved training of wasserstein gans.
\newblock In {\em Advances in Neural Information Processing Systems}, pages
  5769--5779, 2017.

\bibitem{GANRL}
Vincent Huang, Tobias Ley, Martha Vlachou{-}Konchylaki, and Wenfeng Hu.
\newblock Enhanced experience replay generation for efficient reinforcement
  learning.
\newblock {\em CoRR}, abs/1705.08245, 2017.

\bibitem{kingma2013auto}
Diederik~P Kingma and Max Welling.
\newblock Auto-encoding variational bayes.
\newblock {\em arXiv preprint arXiv:1312.6114}, 2013.

\bibitem{krizhevsky2012imagenet}
Alex Krizhevsky, Ilya Sutskever, and Geoffrey~E Hinton.
\newblock Imagenet classification with deep convolutional neural networks.
\newblock In {\em Advances in neural information processing systems}, pages
  1097--1105, 2012.

\bibitem{ddpg}
Timothy~P. Lillicrap, Jonathan~J. Hunt, Alexander Pritzel, Nicolas Heess, Tom
  Erez, Yuval Tassa, David Silver, and Daan Wierstra.
\newblock Continuous control with deep reinforcement learning.
\newblock {\em CoRR}, abs/1509.02971, 2015.

\bibitem{mescheder2018convergence}
Lars Mescheder.
\newblock On the convergence properties of gan training.
\newblock {\em arXiv preprint arXiv:1801.04406}, 2018.

\bibitem{dqn}
Volodymyr Mnih, Koray Kavukcuoglu, David Silver, Alex Graves, Ioannis
  Antonoglou, Daan Wierstra, and Martin~A. Riedmiller.
\newblock Playing atari with deep reinforcement learning.
\newblock {\em CoRR}, abs/1312.5602, 2013.

\bibitem{bootstrappedDQN}
Ian Osband, Charles Blundell, Alexander Pritzel, and Benjamin~Van Roy.
\newblock Deep exploration via bootstrapped {DQN}.
\newblock {\em CoRR}, abs/1602.04621, 2016.

\bibitem{connection_gan_actor_critic}
David Pfau and Oriol Vinyals.
\newblock Connecting generative adversarial networks and actor-critic methods.
\newblock {\em CoRR}, abs/1610.01945, 2016.

\bibitem{rummery1994line}
Gavin~A Rummery and Mahesan Niranjan.
\newblock {\em On-line Q-learning using connectionist systems}, volume~37.
\newblock University of Cambridge, Department of Engineering, 1994.

\bibitem{ruschendorf1985wasserstein}
Ludger R{\"u}schendorf.
\newblock The wasserstein distance and approximation theorems.
\newblock {\em Probability Theory and Related Fields}, 70(1):117--129, 1985.

\bibitem{salakhutdinov2007restricted}
Ruslan Salakhutdinov, Andriy Mnih, and Geoffrey Hinton.
\newblock Restricted boltzmann machines for collaborative filtering.
\newblock In {\em Proceedings of the 24th international conference on Machine
  learning}, pages 791--798. ACM, 2007.

\bibitem{trpo}
John Schulman, Sergey Levine, Philipp Moritz, Michael~I. Jordan, and Pieter
  Abbeel.
\newblock Trust region policy optimization.
\newblock {\em CoRR}, abs/1502.05477, 2015.

\bibitem{ppo}
John Schulman, Filip Wolski, Prafulla Dhariwal, Alec Radford, and Oleg Klimov.
\newblock Proximal policy optimization algorithms.
\newblock {\em CoRR}, abs/1707.06347, 2017.

\bibitem{sutton1988learning}
Richard~S Sutton.
\newblock Learning to predict by the methods of temporal differences.
\newblock {\em Machine learning}, 3(1):9--44, 1988.

\bibitem{tsitsiklis1997analysis}
John~N Tsitsiklis and Benjamin Van~Roy.
\newblock Analysis of temporal-diffference learning with function
  approximation.
\newblock In {\em Advances in neural information processing systems}, pages
  1075--1081, 1997.

\bibitem{watkins1992q}
Christopher~JCH Watkins and Peter Dayan.
\newblock Q-learning.
\newblock {\em Machine learning}, 8(3-4):279--292, 1992.

\end{thebibliography}


\end{document}